\documentclass[lettersize,journal]{IEEEtran}
\usepackage{amsmath,amsfonts}
\usepackage{algorithmic}
\usepackage{algorithm}
\usepackage{array}
\usepackage[caption=false,font=normalsize,labelfont=sf,textfont=sf]{subfig}
\usepackage{textcomp}
\usepackage{stfloats}
\usepackage{url}
\usepackage{verbatim}
\usepackage{graphicx}
\usepackage{cite}
\usepackage{makecell}
\usepackage{multirow}
\usepackage[hidelinks]{hyperref}
\usepackage[table,xcdraw]{xcolor}
\usepackage{booktabs}
\begin{document}

\title{Convolutional Neural Networks and Mixture
of Experts for Intrusion Detection in 5G Networks and beyond}

\author{Loukas Ilias, George Doukas, Vangelis Lamprou, Christos Ntanos, Dimitris Askounis
\thanks{The authors are with the Decision Support Systems Laboratory, School of Electrical and Computer Engineering, National Technical University of Athens, 15780 Athens, Greece (e-mail: lilias@epu.ntua.gr; gdoukas@epu.ntua.gr; vlamprou@epu.ntua.gr; cntanos@epu.ntua.gr; askous@epu.ntua.gr).}
}



\maketitle

\begin{abstract}
The advent of 6G/NextG networks comes along with a series of benefits, including extreme
capacity, reliability, and efficiency. 6G/NextG networks must be equipped with advanced Artificial Intelligence algorithms,
in order to evade new security threats. Existing studies on the intrusion detection task  are based on deep neural networks consisting of static components, which are not
conditional on the input. This limits their representation power and efficiency. To resolve these issues, we
present the first study integrating Mixture of Experts (MoE) for identifying malicious traffic. Specifically,
we use network traffic data and convert the 1D array of features into a 2D matrix. Next, we pass this
matrix through convolutional neural network (CNN) layers followed by batch normalization and max
pooling layers. After this, a sparsely gated MoE layer
is used. This layer consists of a set of experts (dense layers) and a router, where the router assigns weights
to the output of each expert. Sparsity is achieved by choosing the most relevant experts of the total ones.
Finally, we perform a series of ablation experiments to prove the effectiveness of our proposed model.
Experiments are conducted on the 5G-NIDD dataset, a network intrusion detection dataset generated from a
real 5G test network, and the NANCY dataset, which includes cyberattacks on O-RAN 5G Testbed Dataset.
Results show that our introduced approach reaches Accuracy up to 99.96\% and 79.59\% in terms of the
5G-NIDD and NANCY datasets respectively.
Findings also show that our proposed model achieves multiple advantages over state-of-the-art approaches.
\end{abstract}


\section{Introduction}

Fifth generation (5G) networks have found applications in several domains, including autonomous vehicles, smart factories, smart cities, healthcare, and so on, due to their significant improvements in latency, throughput, and badwidth \cite{8869705}. Although, the potential of 5G networks has not fully investigated, both academy and industries have moved their interest towards 6G networks \cite{9482503}. At the same time, several projects are funded under the Smart Networks and Services Joint Undertaking (SNS JU)\footnote{https://smart-networks.europa.eu/} and Horizon 2020 programme \cite{9349624}, in order to facilitate the transition to 6G networks. As new requirements arise in the context of 6G networks, including latency, mobility, peak data rate, spectrum efficiency, area traffic capacity, and network energy efficiency, it must be noted that new attackers also arise \cite{9426946}. Since, these attackers come with advanced characteristics, intelligent threat mitigation systems will be needed. In this context, Artificial Intelligence (AI) can play a pivotal role in protecting 6G networks from attacks, leading to the creation of robust systems.

Existing studies train shallow machine learning classifiers, obtaining in this manner suboptimal performance and poor generalization. Recently, existing studies have moved their interests in converting network traffic data or packet-level data into images and then training convolutional neural networks (CNNs), pretrained CNNs (AlexNet, VGG19, ResNet), RNNs (LSTMs, BiLSTMs), and after that employing fully-connected layers being responsible for classification. However, these approaches depend on dense layers being responsible for doing everything. Network parameters are fixed during training, while inference is performed in a static manner, demanding in this way computational resources and increasing both training and inference time. On the contrary, literature review suggests that models conditioned on the input \cite{9560049} enjoy a series of benefits, including efficiency, representation power, adaptiveness, compatibility. This is attributable to the fact that these models selectively activate model components. Mixture of Experts belong to the category of the input-conditional computation models. Specifically, MoE were originally proposed in \cite{6797059} and since then they have found applications in a series of domains \cite{8100236, cai2024surveymixtureexperts, shazeer2017}.

To address the aforementioned limitations, we present the first study integrating MoE layers into a deep neural network in the task of intrusion detection in 5G networks. Specifically, we use network traffic data consisting of a feature set, which corresponds to a 1D array. We reshape this array into a matrix, which is then given as input to CNN, max-pooling, and batch normalization layers. After this, we use a sparsely-gated MoE layer \cite{shazeer2017}, which applies different subsets of layers (experts) and activates only a selected subset of experts, i.e., the \textit{k} most relevant ones, during each forward pass. Experiments are performed on two publicly available datasets, namely the 5G-NIDD dataset \cite{samarakoon20225g} and the O-RAN 5G Testbed NANCY dataset \cite{liatifis_2024_14811122}. Results demonstrate that the proposed method achieves notable benefits over state-of-the-art approaches.

Our main contributions can be summarized as follows:

\begin{itemize}
    \item To the best of our knowledge, this is the first study employing sparse MoE layers in the intrusion detection task.
    \item We perform our experiments on two publicly available datasets related to 5G networks.
    \item We perform a series of ablation experiments to investigate the effectiveness of the proposed architecture.
\end{itemize}

The rest of this paper is organized as follows: Section~\ref{sec_related_work} presents the existing studies on the intrusion detection task. Section~\ref{sec_dataset} presents the dataset used for conducting our experiments. Section~\ref{sec_methodology} presents the proposed methodology. Section~\ref{sec_experiments_results} presents the experimental setup, results, and the ablation experiments. Finally, Section~\ref{sec_conclusion_future_work} presents some concluding remarks, limitations, and ideas for future work.

\section{Related Work} \label{sec_related_work}

\subsection{Traditional Machine Learning Algorithms}

Kasongo and Sun \cite{kasongo2020performance} trained traditional machine learning algorithms using the UNSW-NB15 dataset. The authors employed XGBoost for selecting the most important features and then they trained Artificial Neural Networks (ANNs), k-Nearest Neighbors (k-NN), Decision Trees (DT), Logistic Regression, and Support Vector Machine (SVM). The authors stated that ANN achieved the highest performance.

In \cite{THAKKAR2023353}, the authors introduced an approach based on feature selection. Specifically, the introduced feature selection strategy is based on the fusion of statistical importance using standard deviation and difference of mean and median. The authors performed their experiments on the NSL-KDD, UNSW-NB15, and CICIDS2017 datasets. The authors compared their introduced feature selection approach with existing approaches, including recursive feature elimination, chi-square, correlation-based feature selection, genetic algorithm, mutual information, Relief-f, and Random Forest. Results showed that the proposed approach outperformed existing ones in all evaluation metrics across all datasets.

\subsection{Deep Neural Networks - CNNs and RNNs}

Hadi et al. \cite{HADI2024124215} presented a multi-tier fusion approach, where several models were trained, including CNNs, GANs, and MLP. Fusion methods, including minimum, maximum, median, sum, and weighted sum, were employed to fuse the outputs of the aforementioned deep learning models. Experiments were conducted on three datasets, including 5G-NIDD, and showed promising results.

Farzaneh et al. \cite{FARZANEH2024107927} introduced three transfer learning strategies for detecting DoS attacks. Specifically, the authors utilized a source and a target dataset. In terms of the source dataset, the authors utilized the dataset introduced in \cite{32k1-dr12-23}, which consists of eight types of DDoS attacks. Regarding the target dataset, the authors utilized the 5G-NIDD dataset. The authors employed CNNs, ResNet, Inception, and BiLSTM. Transfer learning strategies, including freezing some layers and removing the last layer, were employed. Findings showed that the BiLSTM model achieved the best evaluation results.

Sadhwani et al. \cite{sadhwani20245g} used the 5G-NIDD dataset to conduct their experiments. The authors employed variance and correlation reduction followed by a filter-based feature selection approach, namely F1-score, to reduce the dimensionality of the input feature set, keeping at the same time the most informative features. The authors trained and tested a series of shallow machine learning classifiers, including kNN, Naive Bayes, DT, and Random Forest (RF), and deep learning models, including MLP, CNN, LSTM, and CNN-LSTM models. Results showed that CNN-LSTM yielded the highest performance. 

A different approach was introduced by \cite{10634309}, where the authors focused their study on the sequential nature of packets in a network flow. A set of features corresponding to header data was extracted. Next, the authors trained a deep learning model consisting of an LSTM (or GRU) layer followed by an attention mechanism. Experiments were conducted on CICIDS2017 and 5G-NIDD datasets. Results showed that GRU coupled with an attention layer yielded the highest performance in terms of the CICIDS2017 dataset, while LSTM with attention obtained the best results in terms of the 5G-NIDD dataset.

In \cite{LILHORE2024103560}, experiments were performed on CICIDS2017/2018 and UNSW-NB15 datasets. After applying preprocessing techniques and selecting the most relevant features by employing a Decision Tree classifier in combination with Mahalanobis Distance-based Oversampling method, the authors converted data into images. Their proposed approach includes a MobileNet in conjunction with an SVM classifier. As baselines, the authors used VGG16, VGG19, EfficientNet, and InceptionNet. Results showed the strength of the proposed methodology.

A different approach was introduced by \cite{9217754}, where the authors removed socket features and proposed an RNN-based autoencoder. Experiments were performed on the CICDDoS2019 dataset \cite{8888419}. Results demonstrated the effectiveness of the proposed method.

Agrafiotis et al. \cite{10.1145/3590777.3590807} presented a toolkit for converting packets into images. LSTM autoencoders were trained for the generation of embeddings followed by a fully-connected layer for classification. Experiments on 5G-NIDD showed that the model can achieve promising results.

An image-based method, namely MAGNETO, was introduced by \cite{ANDRESINI2021108}. After transforming data into images, the authors trained Generative Adversarial Networks (GANs) for creating new images and thus augmenting the training set. Finally, CNNs were trained on four datasets, namely KDDCUP99\footnote{https://kdd.ics.uci.edu//databases//kddcup99//kddcup99.html}, UNSW-NB15, CICDS2017, and AAGM17\footnote{https://www.unb.ca/cic/datasets/android-adware.html}.

Computer vision approaches were also introduced by \cite{10423385}. Specifically, the authors presented a method used directly on packet-level. A set of features was extracted per packet. After designing the 2D matrix, the authors employed computer vision deep learning models, including Inception, Xception, EfficientNet, MobileNet, DenseNet, ResNet, and a customized CNN. Results showed that the customized CNN yielded the highest results.

Zhang et al. \cite{ZHANG2020107315} introduced a method based on SMOTE and Gaussian Mixture Models for dealing with data imbalance. Finally, the authors trained a deep learning model consisting of CNN layers. The authors compared their method with five class imbalanced processing techniques, including ROS, SMOTE, ADASYN, replacing GMM with RUS and k-means. Results showed that the proposed approach outperformed existing ones.

\subsection{Unsupervised Learning}

A contrastive learning approach was proposed by \cite{10447814}. Specifically, the authors used the 5G-NIDD dataset and an IoT dataset \cite{mirsky2018kitsune}. Next, the authors used both statistical and original (IP, TCP, UDP, payload) features of packets. After that, the authors employed an unsupervised method based on autoencoders, which use contrastive learning. An innovative loss was proposed, which integrates both reconstruction and contrastive loss. Results showed that the proposed approach yielded similar results to existing ones. 

An unsupervised learning approach was introduced by \cite{10287343}. Specifically, the authors used an autoencoder and passed the latent representation vector through a Gaussian Mixture Model. The authors performed their experiments on the CICIDS2017 dataset. Results showed the strength of the proposed approach.

The study in \cite{binbusayyis2021unsupervised} proposed an approach incorporating an one-dimensional convolutional autoencoder and an one-class support vector machine. Experiments were performed on NSL-KDD and UNSW-NB15 datasets. Findings showed the potential of the proposed approach for the design of an effective intrusion detection system.

\subsection{Related Work Review Findings}

Existing studies rely on training traditional machine learning classifiers, thus obtaining suboptimal performance. Image-based methods are employed in conjunction with the development of customized CNNs, pretrained CNNs on the vision domain (AlexNet, Inception, VGG16), and RNNs. However, these methods are fixed during training, perform inference in a static manner, demanding in this way significant computational resources while also increasing training and inference time. 

Our study is different from existing research initiatives, since we present the first study using sparsely gated MoE layers, where only a subset of experts is activated during each forward pass. Additionally, instead of aggregating the outputs from all experts, sparsely gated MoE layers keep only the \textit{k} relevant experts.

\section{Datasets} \label{sec_dataset}

\subsection{5G-NIDD}

We use the 5G-NIDD dataset to conduct our experiments \cite{samarakoon20225g}. This dataset has been collected using the 5G Test Network (5GTN)\footnote{https://5gtnf.fi/}. It contains both pcap files and network traffic data. Contrary to existing datasets, benign traffic have been generated by employing real mobile devices attached to the 5GTN. The advantage of this dataset over existing ones is related to the fact that it contains features existing in 5G network flows. Existing data, including the UNSW-NB15 \cite{7348942} or CICIDS2017/2018 \cite{panigrahi2018detailed, sharafaldin2018toward}, include limitations, since they have been collected in the past in the absence of the technological evolutions. Also, UNSW-NB15 dataset exploits a synthetic environment for generation of attacks.

Data of 5G-NIDD are extracted from two base stations. The authors in \cite{samarakoon20225g} have made available a \textit{csv} file, which corresponds to the combined dataset, consisting of network flow data. 5G-NIDD dataset contains two attack categories, namely DoS/DDoS and Port Scan attacks. In terms of the DoS/DDoS attacks, the following types are included: ICMP flood, UDP flood, SYN flood, HTTP flood, and Slow rate DoS. Regarding the Port Scan attack, the following types are included: SYN Scan, TCP Connect Scan, and UDP Scan. 

\subsubsection{Handling Missing Values}

Several methods have been used for imputing missing values. For instance, the study in \cite{sadhwani20245g} drops some columns, while some missing values are imputed using the mean of the respective columns. However, dropping columns results in loss of information. In \cite{10492991}, the authors impute missing data using median or zero. 

In this study, missing values in numerical features are imputed using the mean value per class, while categorical features are imputed with the most frequent value per class.

\subsubsection{Designing the feature set}

In Table~\ref{feature_analysis}, we describe the features used in our experiments. To normalize numerical features, Min-Max Scaling is applied, ensuring values fall between 0 and 1. Categorical features are one-hot encoded, excluding the first level to avoid redundancy.

In total, we use 78 features. This 1D array is transformed into a 2D matrix, i.e., $6\times13$, and is given as input to CNN layers, which are described in detail in the next section.

\begin{table*}[!htb]
\centering
\caption{Description of features used (5G-NIDD dataset)}
\renewcommand{\arraystretch}{1.2}
\resizebox{0.8\textwidth}{!}{ 
\begin{tabular}{p{2cm}|p{4cm}|p{3cm}|p{4cm}} 
\toprule
\rowcolor[HTML]{EFEFEF} 
\textbf{Features} & \textbf{Description} & \textbf{Type} & \textbf{No. of features after preprocessing} \\ \hline
\textbf{Seq} & Sequence Number & Numerical & 1 \\ \hline
\textbf{Dur} & Duration & Numerical & 1 \\ \hline
\textbf{RunTime} & Run Time & Numerical & 1 \\ \hline
\textbf{Mean} & Mean Value & Numerical & 1 \\ \hline
\textbf{Sum} & Sum Value & Numerical & 1 \\ \hline
\textbf{Min} & Minimum Value & Numerical & 1 \\ \hline
\textbf{Max} & Maximum Value & Numerical & 1 \\ \hline
\textbf{Proto} & Protocol Type & Categorical & 7 \\ \hline
\textbf{sTos} & \multirow{2}{*}{Type of Service values} & Numerical & 1 \\ \cline{1-1} \cline{3-4} 
\textbf{dTos} &  & Numerical & 1 \\ \hline
\textbf{sDSb} & \multirow{2}{*}{\makecell[l]{Source and Destination \\ Behavioral Flags}} & Categorical & 11 \\ \cline{1-1} \cline{3-4} 
\textbf{dDSb} &  & Categorical & 5 \\ \hline
\textbf{sTtl} & \multirow{2}{*}{\makecell[l]{Time-to-Live values for source \\ and destination}} & Numerical & 1 \\ \cline{1-1} \cline{3-4} 
\textbf{dTtl} &  & Numerical & 1 \\ \hline
\textbf{sHops} & \multirow{2}{*}{\makecell[l]{Number of hops taken by packets \\ from source to destination}} & Numerical & 1 \\ \cline{1-1} \cline{3-4} 
\textbf{dHops} &  & Numerical & 1 \\ \hline
\textbf{Cause} & Cause code & Categorical & 2 \\ \hline
\textbf{TotPkts} & Total Packets & Numerical & 1 \\ \hline
\textbf{SrcPkts} & Source Packets & Numerical & 1 \\ \hline
\textbf{DstPkts} & Destination Packets & Numerical & 1 \\ \hline
\textbf{TotBytes} & Total Bytes & Numerical & 1 \\ \hline
\textbf{SrcBytes} & Source Bytes & Numerical & 1 \\ \hline
\textbf{DstBytes} & Destination Bytes & Numerical & 1 \\ \hline
\textbf{Offset} & Offset Value & Numerical & 1 \\ \hline
\textbf{sMeanPktSz} & Source Mean Packet Size & Numerical & 1 \\ \hline
\textbf{dMeanPktSz} & Destination Mean Packet Size & Numerical & 1 \\ \hline
\textbf{Load} & Load Value & Numerical & 1 \\ \hline
\textbf{SrcLoad} & Source Load & Numerical & 1 \\ \hline
\textbf{DstLoad} & Destination Load & Numerical & 1 \\ \hline
\textbf{Loss} & Total Loss & Numerical & 1 \\ \hline
\textbf{SrcLoss} & Source Loss & Numerical & 1 \\ \hline
\textbf{DstLoss} & Destination Loss & Numerical & 1 \\ \hline
\textbf{pLoss} & Packet Loss & Numerical & 1 \\ \hline
\textbf{SrcGap} & \multirow{2}{*}{Time Gaps} & Numerical & 1 \\ \cline{1-1} \cline{3-4} 
\textbf{DstGap} &  & Numerical & 1 \\ \hline
\textbf{Rate} & \multirow{3}{*}{Transmission Rates} & Numerical & 1 \\ \cline{1-1} \cline{3-4} 
\textbf{SrcRate} &  & Numerical & 1 \\ \cline{1-1} \cline{3-4} 
\textbf{DstRate} &  & Numerical & 1 \\ \hline
\textbf{State} & State Information & Categorical & 10 \\ \hline
\textbf{SrcWin} & \multirow{2}{*}{TCP Window Sizes} & Numerical & 1 \\ \cline{1-1} \cline{3-4} 
\textbf{DstWin} &  & Numerical & 1 \\ \hline
\textbf{sVid} & \multirow{2}{*}{VLAN IDs} & Numerical & 1 \\ \cline{1-1} \cline{3-4} 
\textbf{dVid} &  & Numerical & 1 \\ \hline
\textbf{SrcTCPBase} & \multirow{2}{*}{TCP Base Values} & Numerical & 1 \\ \cline{1-1} \cline{3-4} 
\textbf{DstTCPBase} &  & Numerical & 1 \\ \hline
\textbf{TcpRtt} & TCP Round Trip-Time & Numerical & 1 \\ \hline
\textbf{SynAck} & \multirow{2}{*}{Specific Packet Types} & Numerical & 1 \\ \cline{1-1} \cline{3-4} 
\textbf{AckDat} &  & Numerical & 1 \\ \midrule
\textbf{Total} & - & - & \textbf{78} \\ \bottomrule
\end{tabular}
}
\label{feature_analysis}
\end{table*}

\subsection{NANCY}

NANCY dataset is a newly collected dataset. For constructing this dataset, a malicious user carries out cyberattacks against various services running in the main operator and the micro-operator. This dataset contains the following types of attacks, Reconnaissance Attack, TCP Scan, SYN Scan, SYN Flood, HTTP Flood, and Slow-rate DoS. The authors have used CICFlowMeter\footnote{https://www.unb.ca/cic/research/applications.html} software for obtaining the network flows. Due to data imbalance issues, we remove the Slowrate DoS and Reconnaissance Attack from our dataset. Specifically, the NANCY dataset includes 3,553 samples correponding to Slowrate DoS and 2,044 samples corresponding to Reconnaissance Attack. 

Contrary to the 5G-NIDD dataset, the NANCY dataset does not include any missing values.

\paragraph{Designing the feature set} We keep only the numerical features. The complete feature set is reported in Table~\ref{tab:cicflowmeter-features-bold}. In total, we use 72 features. Min-Max Scaling is applied, ensuring values fall between 0 and 1. This 1D array is transformed into a 2D matrix, i.e., $6 \times 12$, and is given as input to CNN layers, which are described in detail in Section~\ref{sec_methodology}.

\begin{table*}[htbp]
\centering
\caption{Description of Features used (NANCY dataset)}
\begin{tabular}{l|p{5cm}|l|p{5cm}}
\toprule
\rowcolor[HTML]{EFEFEF} 
\textbf{Feature} & \textbf{Description} & \textbf{Feature} & \textbf{Description} \\
\midrule
\textbf{Flow Duration} & Duration of the flow in microseconds & \textbf{Tot Fwd Pkts} & Total number of packets in the forward direction \\
\textbf{Tot Bwd Pkts} & Total number of packets in the backward direction & \textbf{TotLen Fwd Pkts} & Total length of packets in the forward direction \\
\textbf{TotLen Bwd Pkts} & Total size of packets in the backward direction & \textbf{Fwd Pkt Len Max} & Maximum size of forward packets \\
\textbf{Fwd Pkt Len Min} & Minimum size of forward packets & \textbf{Fwd Pkt Len Mean} & Mean size of forward packets \\
\textbf{Fwd Pkt Len Std} & Standard deviation size of packet in forward direction & \textbf{Bwd Pkt Len Max} & Maximum size of packets in backward direction \\
\textbf{Bwd Pkt Len Min} & Minimum size of backward packets & \textbf{Bwd Pkt Len Mean} & Mean size of backward packets \\
\textbf{Bwd Pkt Len Std} & Standard deviation size of packet in backward direction & \textbf{Flow Byts/s} & Number of flow bytes per second \\
\textbf{Flow Pkts/s} & Number of flow packets per second & \textbf{Flow IAT Mean} & Mean time between two packets sent in the flow \\
\textbf{Flow IAT Std} & Standard deviation time between two packets sent in the flow & \textbf{Flow IAT Max} & Maximum time between two packets sent in the flow \\
\textbf{Flow IAT Min} & Minimum time between two packets sent in the flow & \textbf{Fwd IAT Tot} & Total time between two packets sent in the forward direction \\
\textbf{Fwd IAT Mean} & Mean time between two packets sent in the forward direction & \textbf{Fwd IAT Std} & Standard deviation time between two packets sent in the forward direction \\
\textbf{Fwd IAT Max} & Maximum time between two packets sent in the forward direction & \textbf{Fwd IAT Min} & Minimum time between two packets sent in the forward direction \\
\textbf{Bwd IAT Tot} & Total time between two packets sent in the backward direction & \textbf{Bwd IAT Mean} & Mean time between two packets sent in the backward direction \\
\textbf{Bwd IAT Std} & Standard deviation time between two packets sent in the backward
direction & \textbf{Bwd IAT Max} & Maximum time between two packets sent in the backward direction \\
\textbf{Bwd IAT Min} & Minimum time between two packets sent in the backward direction & \textbf{Fwd Header Len} & Total header length in forward direction \\
\textbf{Bwd Header Len} & Total header length in backward direction & \textbf{Fwd Pkts/s} & Number of forward packets per second \\
\textbf{Bwd Pkts/s} & Number of backward packets per second & \textbf{Pkt Len Min} & Minimum packet length \\
\textbf{Pkt Len Max} & Maximum packet length & \textbf{Pkt Len Mean} & Mean packet length \\
\textbf{Pkt Len Std} & Standard deviation of packet lengths & \textbf{Pkt Len Var} & Variance of packet lengths \\
\textbf{FIN Flag Cnt} & Number of packets with FIN flag & \textbf{SYN Flag Cnt} & Number of packets with SYN flag \\
\textbf{RST Flag Cnt} & Number of packets with RST flag & \textbf{PSH Flag Cnt} & Number of packets with PSH flag \\
\textbf{ACK Flag Cnt} & Number of packets with ACK flag & \textbf{URG Flag Cnt} & Number of packets with URG flag \\
\textbf{CWE Flag Count} & Number of CWE flag packets & \textbf{ECE Flag Cnt} & Number of packets with ECE flag \\
\textbf{Down/Up Ratio} & Ratio of bytes sent in backward and forward directions & \textbf{Pkt Size Avg} & Average packet size \\
\textbf{Fwd Seg Size Avg} & Average segment size in forward direction & \textbf{Bwd Seg Size Avg} & Average segment size in backward direction \\
\textbf{Fwd Byts/b Avg} & Average number of bytes per bulk in forward direction & \textbf{Fwd Pkts/b Avg} & Average number of packets per bulk in forward direction \\
\textbf{Fwd Blk Rate Avg} & Average bulk rate in forward direction & \textbf{Bwd Byts/b Avg} & Average bytes per bulk in backward direction \\
\textbf{Bwd Pkts/b Avg} & Average packets per bulk in backward direction & \textbf{Bwd Blk Rate Avg} & Average bulk rate in backward direction \\
\textbf{Subflow Fwd Pkts} & Total packets in forward subflows & \textbf{Subflow Fwd Byts} & Total bytes in forward subflows \\
\textbf{Subflow Bwd Pkts} & Total packets in backward subflows & \textbf{Subflow Bwd Byts} & Total bytes in backward subflows \\
\textbf{Init Fwd Win Byts} & Initial window bytes in forward direction & \textbf{Init Bwd Win Byts} & Initial window bytes in backward direction \\
\textbf{Fwd Act Data Pkts} & Number of forward active data packets & \textbf{Fwd Seg Size Min} & Minimum segment size in forward direction \\
\textbf{Active Mean} & Mean active time & \textbf{Active Std} & Std of active time \\
\textbf{Active Max} & Maximum activity time & \textbf{Active Min} & Minimum activity time \\
\textbf{Idle Mean} & Mean idle time & \textbf{Idle Std} & Std of idle time \\
\textbf{Idle Max} & Maximum idle time & \textbf{Idle Min} & Minimum idle time \\
\bottomrule
\end{tabular}
\label{tab:cicflowmeter-features-bold}
\end{table*}

\section{Methodology} \label{sec_methodology}

\begin{figure*}[!htb]
    \centering
    \includegraphics[width=0.8\textwidth]{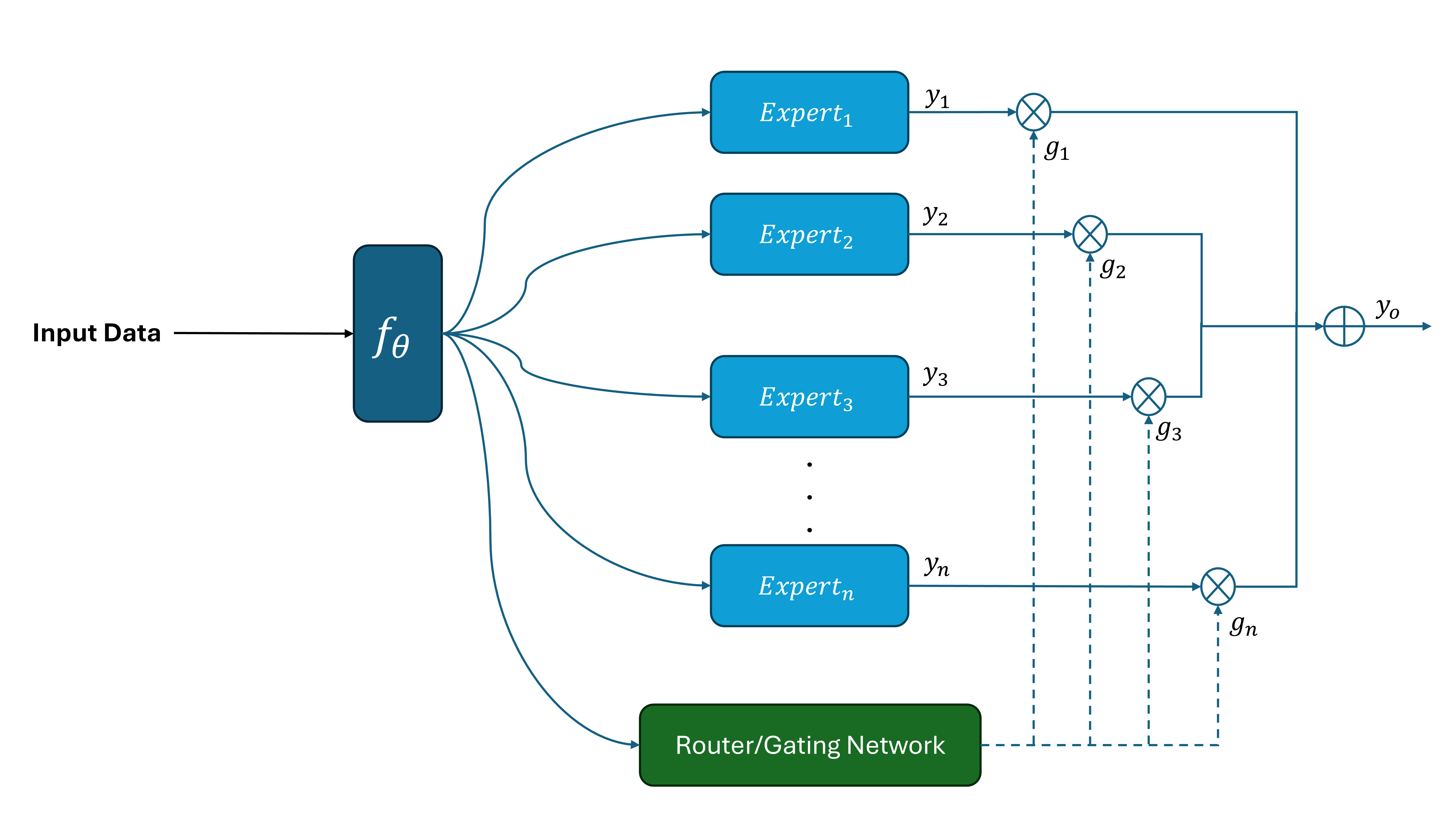}
    \caption{Proposed Methodology}
    \label{proposed_methodology}
    \end{figure*}

\begin{figure}
    \centering
    \subfloat[CNN Architecture ($f_{\theta}$)\label{cnn_architecture}]{
        \includegraphics[width=0.95\columnwidth]{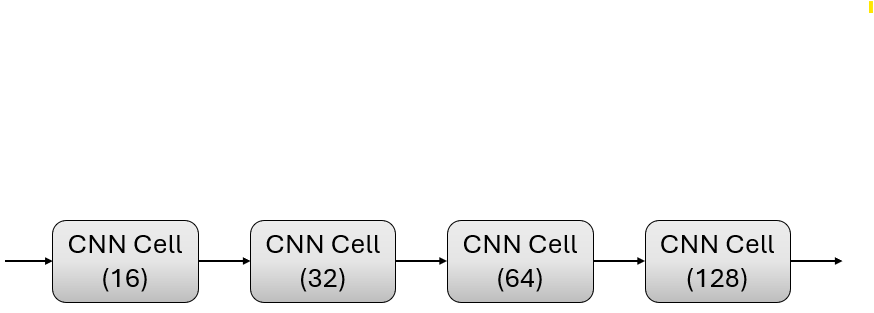}} 

    \subfloat[CNN cell structure\label{cnn_cell_structure}]{\includegraphics[width=0.8\columnwidth]{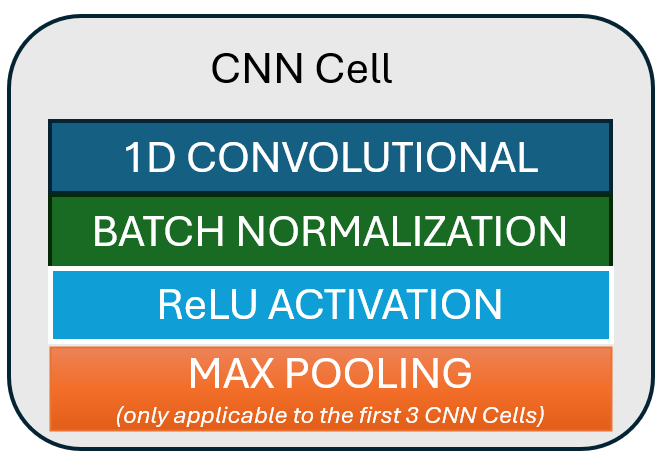}}
    
    \caption{Description of $f_\theta$}
    \label{entire_methodology}
\end{figure}

Our proposed architecture is illustrated in Figures~\ref{proposed_methodology} and \ref{entire_methodology}. Below, we explain in detail each component of our proposed architecture.

\subsection{Convolutional Neural Networks}

We design a convolutional neural network architecture to obtain a representation vector from input $X$. The CNN component of our methodology is illustrated in Fig.~\ref{cnn_architecture}. 

The structure of the CNN cell is illustrated in Fig.~\ref{cnn_cell_structure}. We place 4 CNN cell in a row with 16, 32, 64, and 128 filters respectively. Kernel size and padding are set equal to 3 and 1 respectively. In terms of the max pooling layer, we set both kernel size and stride equal to 2. Each CNN cell, except for the final one consists of an 1D Convolutional, Batch normalization, ReLU activation, and Max-Pooling layer. The final CNN cell does not include a max pooling layer. 

Let the output of this model's component be $x=f_\theta (X) \in \mathbb{R}^d$, where $d=128$.

\subsection{Sparse Mixture of Experts}

In this section, we describe the sparsely-gated MoE layer. Specifically, this layer consists of \textit{n} experts and a router/gating network. The aim of the router is to determine the important experts per input data. 

Eq.~\ref{moe_output} shows the output of the MoE layer, denoted as $y_o$. $y_i$ denotes the output of each expert, while $g_i$ indicates the weight assigned by the router to each expert's output $y_i$. Each expert consists of a deep neural network with a hidden layer of 16 units.

\begin{equation}
    y_o  = \sum_{i=1}^n g_i (x) y_i (x) 
    \label{moe_output}
\end{equation}

Equations~\ref{first_equation}-\ref{third_equation} describe the sparsity mechanism and the addition of noise. Specifically, $TopK(\cdot, k)$ function selects the $k$ most relevant experts by setting the output vectors at their original values and setting all the other values to $-\infty$. Values of $-\infty$ become close to zero after applying a $softmax$ function. The addition of noise term, denoted as $R_{noise}$ facilitates load balancing among experts (see below $L_{load}$) and enhances the stability of MoE training.

\begin{equation}
    \mathcal{G}(\mathbf{x}; \boldsymbol{\Theta})_i = \text{softmax}(\text{TopK}(g(\mathbf{x}; \boldsymbol{\Theta}) + \mathcal{R}_{\text{noise}}, k))_i,  
    \label{first_equation}
\end{equation}

, where $\mathcal{R}_{\text{noise}} = StandardNormal()\cdot Softplus((x\cdot W_{noise})_i)$

\begin{equation}
    \text{TopK}(g(\mathbf{x}; \boldsymbol{\Theta}), k)_i = 
    \begin{cases} 
      g(\mathbf{x}; \boldsymbol{\Theta})_i, & \text{condition,} \\
      -\infty, & \text{otherwise.}
    \end{cases} 
    \label{topk_equations}
\end{equation}

, where:
\begin{equation}
    \text{condition : } \text{if } g(\mathbf{x}; \boldsymbol{\Theta})_i \text{ is in the top-$k$ elements of } g(\mathbf{x}; \boldsymbol{\Theta}). 
    \label{third_equation}
\end{equation}

For load-balancing purposes and balancing expert utilization, we design two losses based on the study of \cite{shazeer2017}.

\begin{itemize}

\item \textbf{$\mathcal{L}_{importance}$}: \cite{shazeer2017} observe that large weights are assigned to specific experts by the router. To adddress this issue and ensure uniform routing weights across all experts, the authors in \cite{shazeer2017} design the following loss function. This loss aims at equal importance across all experts.

\begin{equation}
    \mathcal{L}_{\text{importance}} = w_{importance} \cdot CV(Importance(X))^2
\end{equation}

, where $X$ is the batch of features, $CV(\cdot) = \frac{Std(\cdot)}{Mean(\cdot)}$ and $Importance(\cdot)$ is given by the equation below:

\begin{equation}
    Importance(X) = \sum_{x \in X} G(x)
\end{equation}

\item \textbf{$\mathcal{L}_{load}$}: \cite{shazeer2017} design the following loss function, in order to ensure that all experts receive equal number of training instances.

\begin{equation}
\small
    P(x,i) = \Phi \left(\frac{\left(x\cdot W_g  \right)_i - kth\_excluding \left(H(x),k,i \right)}{Softplus((x\cdot W_{noise})_i)}  \right)
\end{equation}

, where $\Phi$ is the cumulative distribution function of the standard normal distribution.

\begin{equation}
    Load(X)_i = \sum_{x \in X} P(x,i)
\end{equation}

, where $Load(X)_i$ is the load of the \textit{i}th expert.

\begin{equation}
    \mathcal{L}_{load} (X) = w_{load} \cdot CV (Load(X))^2
\end{equation}

\end{itemize}

\subsection{Loss Function}

We minimize the following loss function:

\begin{equation}
    \mathcal{L} = \mathcal{L}_{cross-entropy} + \alpha \cdot (\mathcal{L}_{importance} + \mathcal{L}_{load})
    \label{alpha_equation}
\end{equation}
, where $\mathcal{L}_{cross-entropy}$ corresponds to the cross-entropy loss function and $\alpha$ is a hyperparameter denoting the importance we place on these two loss functions.

\section{Experiments and Results} \label{sec_experiments_results}

\subsection{Baselines}

In terms of the 5G-NIDD dataset, we compare our method with the following research studies:
\begin{itemize}
    \item Embeddings \& FC (multi-class) \cite{10.1145/3590777.3590807}: This method trains LSMT autoencoders followed by a fully-connected layer. 
    \item FC Sehan (multi-class) \cite{samarakoon20225g}: This method trains a Multilayer Perceptron.
    \item Customized CNN (N=100) \cite{10423385}: This method extracts a set of features per packet, converts flows into an image, and trains a CNN followed by ReLU activated fully-connected layers.
    \item CNN-LSTM (multi-class) \cite{sadhwani20245g}: This method trains a deep neural network consisting of CNN, LSTM, and fully-connected layers.
    \item Fusion Multi-Tier DNN \cite{HADI2024124215}: This method proposes a multi-tier fusion approach, where multiple deep learning models, including MLP, CNNs, and GANs, are trained with the final classification obtained through some combination rules, including maximum, minimum, median, sum, and weighted sum.
\end{itemize}

Since NANCY dataset is a newly collected dataset, there are not research studies performing their experiments on this dataset.

\subsection{Experimental Setup}
We use PyTorch \cite{10.5555/3454287.3455008} for performing our experiments. All experiments are performed on a NVIDIA A100 80GB PCIe GPU.
\paragraph{5G-NIDD}

We use 128 (value of $n$) experts. We keep the 32 (value of $k$) most relevant experts. We set $\alpha$ of Eq.~\ref{alpha_equation} equal to 0.1. We use a batch size of 1024. We divide the dataset into a train and test set (60-40\%) in a stratified manner. We train the models for a maximum of 40 epochs. 

\paragraph{NANCY}

We use 64 (value of $n$) experts. We keep the 32 (value of $k$) most relevant experts. We set $\alpha$ of Eq.~\ref{alpha_equation} equal to 0.1. We use a batch size of 256. Also, we apply class weights. We divide the dataset into a train and test set (70-30\%) in a stratified manner. We train the models for a maximum of 40 epochs.


\subsection{Evaluation Metrics}

Precision, Recall, and F1-score are reported for each class. Accuracy is also reported. 

Regarding 5G-NIDD, we also report weighted F1-score, since we have used an imbalanced dataset.

\subsection{Results}

\paragraph{Results on the 5G-NIDD dataset} Results of our introduced methodology are reported in Table~\ref{results_ourapproach}. Specifically, this table reports the results, i.e., Precision, Recall, and F1-score, per class and the overall Accuracy. We observe that our introduced model achieves F1-score greater than 0.99900 in terms of the Benign, TCP Connect Scan, ICMP flood, and UDP flood classes. Specifically, attacks of ICMP flood and UDP flood are perfectly recognized by our model achieving 1 across all metrics. F1-score greater than (or equal to) 0.99800 (but lower than 0.99900) is yielded in terms of SYN Scan, SYN flood, and HTTP flood. Finally, we observe that our model performs not so well in recognizing UDP Scan and Slow rate DoS reaching F1-score up to 0.99796 and 0.99735 respectively.

\begin{table}[!htb]
\centering
\caption{Results of our proposed approach (5G-NIDD)}
\begin{tabular}{l|c|c|c|c}
\toprule
\rowcolor[HTML]{EFEFEF} 
\textbf{Class}      & \textbf{Accuracy} & \textbf{Precision} & \textbf{Recall} & \textbf{F1 Score} \\ 
\midrule
Benign                  & \multirow{9}{*}{0.99958} & 0.99996                  & 0.99993               & 0.99995                 \\ 
SYN Scan                  &                          & 0.99888                  & 0.99713               & 0.99800                 \\ 
TCP Connect Scan          &                          & 0.99975                  & 0.99950               & 0.99963                 \\ 
UDP Scan                  &                          & 0.99702                  & 0.99890               & 0.99796                 \\ 
ICMP flood                &                          & 1.0                  & 1.0               & 1.0                 \\ 
UDP flood                 &                          & 1.0                  & 1.0               & 1.0                 \\ 
SYN flood                 &                          & 0.99846                  & 0.99949               & 0.99897                 \\ 
HTTP flood                &                          & 0.99793                  & 0.99922               & 0.99857                 \\ 
Slow rate DoS             &                          & 0.99849                  & 0.99621               & 0.99735                 \\ 
\bottomrule
\end{tabular}
\label{results_ourapproach}
\end{table}

Table~\ref{performance_comparison} reports results of our method in comparison with state-of-the-art approaches. As one can observe, our proposed model outperforms existing studies in Precision, Recall, and Accuracy by 0.00044-0.00875, 0.00043-0.01577, and 0.00148-0.00835 respectively. Our method outperforms, also, existing studies apart from \textit{Customized CNN (N=100) \cite{10423385}} in F1-score by 0.01228-0.00194. However, it must be noted that the study of \cite{10423385} outperforms our study by a small margin of 0.00001. In the case of our study, we also report weighted F1-score, in order to take data imbalance into account. As one can observe, our proposed model reaches weighted F1-score up to 0.99958. Therefore, our study considering conditional-input components enjoys the benefits of lower training and inference time among else and achieves comparable performance to state-of-the-art results.

\begin{table*}[htbp]
\centering
\caption{Performance comparison among proposed models and baselines on the 5G-NIDD dataset}
\begin{tabular}{lccccc}
\toprule
\multicolumn{1}{l}{}&\multicolumn{5}{c}{\textbf{Evaluation metrics}}\\
\cline{2-6} 
\multicolumn{1}{l}{\textbf{Architecture}}&\textbf{Precision}&\textbf{Recall}&\textbf{F1-score}&\textbf{Accuracy}&\textbf{Weighted F1-score}\\
\midrule
\multicolumn{6}{>{\columncolor[gray]{.8}}l}{\textbf{Comparison with state-of-the-art}} \\
\textit{Embeddings \& FC (multi-class) \cite{10.1145/3590777.3590807}} & 0.99019 & 0.98316 & 0.98666 & 0.99123 & - \\ \hline
\textit{FC Sehan \cite{samarakoon20225g}} & 0.99167 & 0.98869 & 0.99017 & 0.99499 & - \\ \hline
\textit{Customized CNN (N=100) \cite{10423385}} & - & - & 0.99895 & - & - \\ \hline
\textit{CNN-LSTM (multi-class) \cite{sadhwani20245g}} & 0.99850 & 0.99850 & 0.99700 & 0.99810 & - \\ \hline
\textit{Fusion Multi-Tier DNN \cite{HADI2024124215}}  & - & - & - & 0.99150 & - \\
\midrule
\multicolumn{6}{>{\columncolor[gray]{.8}}l}{\textbf{Introduced Approach}} \\
& 0.99894 & 0.99893 & 0.99894 & 0.99958 & 0.99958 \\ 
\bottomrule
\end{tabular}
\label{performance_comparison}
\end{table*}

\paragraph{Results on the NANCY dataset} 

Table~\ref{results_ourapproach_nancy} presents the performance metrics of the proposed approach on the NANCY dataset, including Accuracy, Precision, Recall, and F1 Score for five traffic classes: Benign, SYN Scan, TCP Connect Scan, SYN flood, and HTTP flood. The overall accuracy of the model across all classes is 0.79592, indicating that approximately 79.6\% of instances were correctly classified. Among the individual classes, the benign class achieved high performance, with a precision of 0.99637, recall of 0.97739, and F1 score of 0.98679, suggesting that normal traffic was accurately detected with minimal misclassifications.

In contrast, the model struggled to effectively detect SYN Scan traffic, achieving the lowest performance metrics with a precision of 0.59807, recall of 0.39572, and F1 score of 0.47629. This indicates a significant number of false negatives, pointing to challenges in identifying this type of scan-based intrusion. TCP Connect Scan performed moderately better, with a precision of 0.64581, recall of 0.74404, and an F1 score of 0.62969, reflecting a more balanced but still limited detection capability.

On the other hand, the approach showed excellent performance in detecting flooding attacks. SYN flood and HTTP flood traffic were classified with high precision and recall values—0.99865 and 0.98469 for SYN flood, and 0.99787 and 0.99760 for HTTP flood, respectively—resulting in F1 scores of 0.99162 and 0.99774. These results demonstrate the model’s robustness in identifying high-volume, aggressive attacks, while also highlighting the need for improvement in detecting subtler, low-rate intrusion attempts such as SYN scans.

\begin{table}[!htb]
\centering
\caption{Results of our proposed approach on the NANCY dataset}
\begin{tabular}{l|c|c|c|c}
\toprule
\rowcolor[HTML]{EFEFEF} 
\textbf{Class} & \textbf{Accuracy} & \textbf{Precision} & \textbf{Recall} & \textbf{F1 Score} \\ 
\midrule
Benign              & \multirow{5}{*}{0.79592} & 0.99637 & 0.97739 & 0.98679 \\ 
SYN Scan            &                   & 0.59807 & 0.39572 & 0.47629 \\ 
TCP Connect Scan    &                   & 0.64581 & 0.74404 & 0.62969 \\ 
SYN flood           &                   & 0.99865 & 0.98469 & 0.99162 \\ 
HTTP flood          &                   & 0.99787 & 0.99760 & 0.99774 \\ 
\bottomrule
\end{tabular}
\label{results_ourapproach_nancy}
\end{table}

\subsection{Ablation Study} \label{sec_ablation_study}

In this section, we perform a series of ablation experiments to explore the effectiveness of the proposed architecture. Results of the ablation study are reported in Table~\ref{Ablation_Study}. 

\paragraph{5G-NIDD}
Firstly, we set $L_{load}=L_{importance}=0$. Results showed a decrease in terms of Accuracy by 0.00103. Specifically, F1-scores in terms of the Benign, SYN Scan, UDP Scan, and HTTP flood presented a decrease of 0.00002, 0.0005, 0.00086, and 0.00431 respectively. By setting both losses equal to zero, specific experts receive the most training instances, while larger weights are assigned to specific experts by the router. This fact deteriorates performance of the MoE layer. 

Secondly, we remove the MoE layer and use a dense layer instead. To be more specific, the output vector of the CNN layers ($128D$) is passed through a dense layer consisting of 9 units (output layer). Results showed that Accuracy presented a decrease from 0.99958 to 0.99829. Specifically, F1-score of the Benign, SYN Scan, ICMP flood, SYN flood, HTTP flood, and Slow rate DoS presented a decrease of 0.00003, 0.00069, 0.00216, 0.00026, 0.00532, and 0.01052 respectively. 

Finally, we remove both CNN and MoE layers, and train a deep neural network consisting of an input layer (78 units) and an output layer (9 units). Findings showed that Accuracy was decreased by 0.00802. F1-scores of all the classes, except for SYN Scan, presented a decrease. Specifically, F1-score of Benign, TCP Connect Scan, UDP Scan, ICMP flood, UDP flood, SYN flood, HTTP flood, and Slow rate DoS presented a decrease of 0.00515, 0.00082, 0.00063, 0.00324, 0.00001, 0.0123, 0.02683, and 0.0451 respectively.

\begin{table}[htbp]
\centering
\caption{Ablation Study (5G-NIDD dataset)}
\begin{tabular}{lcccc}
\toprule
\multicolumn{1}{l}{}&\multicolumn{4}{c}{\textbf{Evaluation metrics}}\\
\cline{2-5} 
\multicolumn{1}{l}{\textbf{Architecture}}&\textbf{Accuracy}&\textbf{Precision}&\textbf{Recall}&\textbf{F1-score}\\
\midrule
\multicolumn{5}{>{\columncolor[gray]{.8}}l}{\textbf{Set importance and load loss to zero}} \\
Benign                  & \multirow{9}{*}{0.99855} & 0.99996                  &    0.99989            &   0.99993               \\ 
SYN Scan                  &                          &   0.99925                &        0.99576        &     0.99750             \\ 
TCP Connect Scan          &                          &   0.99988                &      0.99950          &       0.99969           \\ 
UDP Scan                  &                          &   0.99484                &        0.99937        &     0.99710             \\ 
ICMP flood                &                          &  1.0                  &       1.0         &       1.0           \\ 
UDP flood                 &                          &  1.0                 &      1.0          &       1.0           \\ 
SYN flood                 &                          &  0.99846                 &      0.99949          &     0.99897             \\ 
HTTP flood                &                          &     0.99113              &        0.99741        &     0.99426             \\ 
Slow rate DoS   &                          &     0.99485              &        0.98311        &     0.98894             \\ 
\midrule
\multicolumn{5}{>{\columncolor[gray]{.8}}l}{\textbf{Remove MoE layer}} \\
Benign                  & \multirow{9}{*}{0.99829} &  0.99996                 &      0.99987          &    0.99992              \\ 
SYN Scan                  &                          &  0.99925                 &      0.99538          &      0.99731            \\ 
TCP Connect Scan          &                          &  0.99975                 &    0.99950            &    0.99963              \\ 
UDP Scan                  &                          &   0.99452                &    0.99921            &  0.99686                \\ 
ICMP flood                &                          &   0.99569                &   1.0             &     0.99784             \\ 
UDP flood                 &                          &    1.0               &        1.0        &        1.0          \\ 
SYN flood                 &                          &  0.99795                 &     0.99949           &        0.99871          \\ 
HTTP flood                &                          &    0.98819               &     0.99837           &    0.99325              \\ 
Slow rate DoS             &                          &    0.99658               &      0.97726          &       0.98683           \\ 
\midrule
\multicolumn{5}{>{\columncolor[gray]{.8}}l}{\textbf{Simple input and dense layer}} \\
Benign                  & \multirow{9}{*}{0.99156} &       0.99996            &    0.98970            &    0.99480              \\ 
SYN Scan                  &                          &   0.99888                &      0.99738          &     0.99813             \\ 
TCP Connect Scan          &                          &   0.99988                &     0.99776           &    0.99881              \\ 
UDP Scan                  &                          &   0.99639                &       0.99827         &        0.99733          \\ 
ICMP flood                &                          &    0.99355               &      1.0          &       0.99676           \\ 
UDP flood                 &                          &   0.99997                &       1.0         &        0.99999          \\ 
SYN flood                 &                          &   0.97418                &      0.99949          &        0.98667          \\ 
HTTP flood                &                          &   0.95971                &         0.98407       &       0.97174           \\ 
Slow rate DoS             &                          &   0.94523                &         0.95938       &        0.95225          \\ 

\bottomrule
\end{tabular}
\label{Ablation_Study}
\end{table}

\begin{figure}
    \centering
    \includegraphics[width=1.1\linewidth]{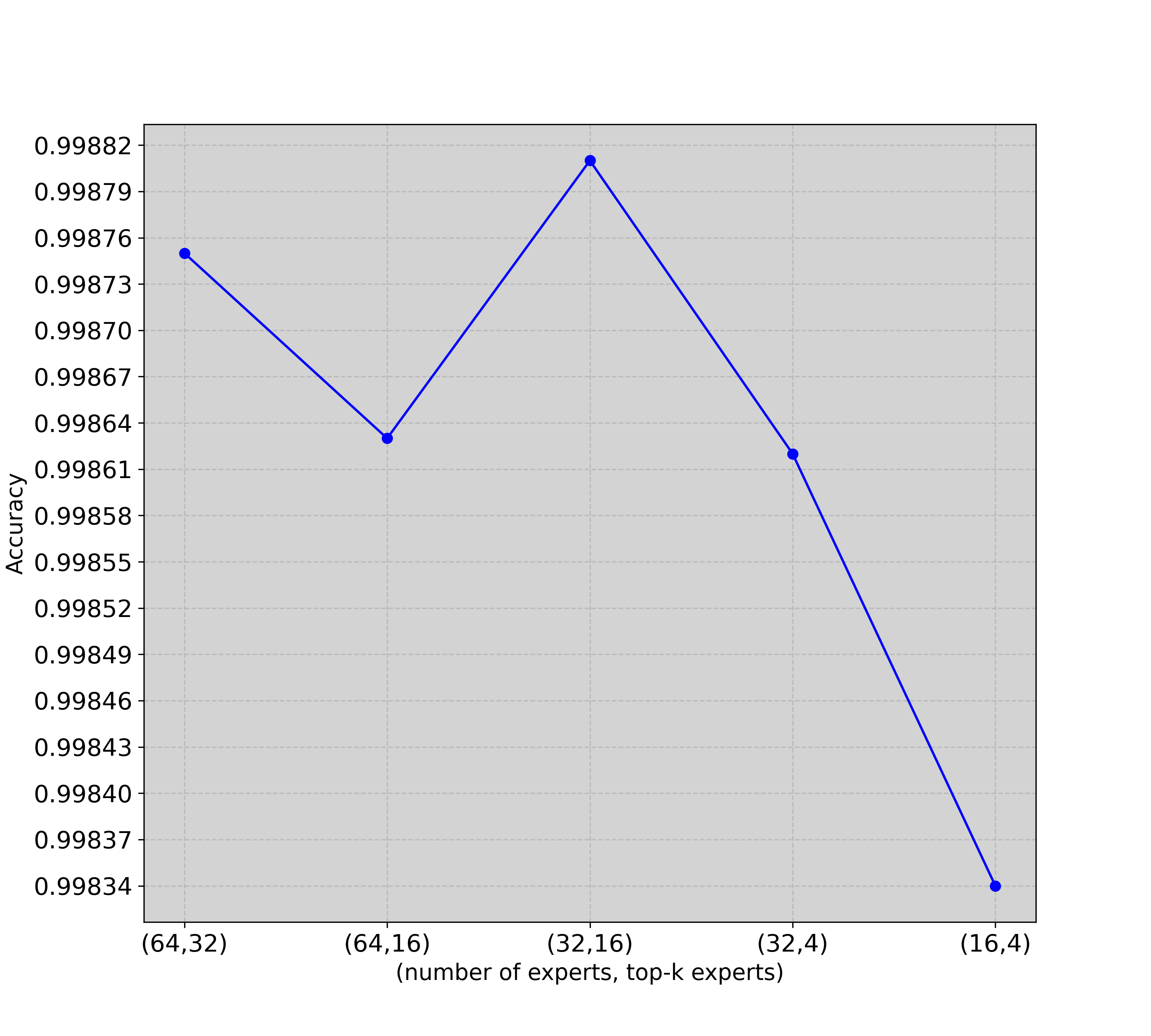}
    \caption{Ablation Study (5G-NIDD dataset). Varying the number of experts and top-\textit{k} experts.}
    \label{ablation_study_experts_kexperts}
\end{figure}

Finally, in Fig.~\ref{ablation_study_experts_kexperts}, we vary the number of experts ($n$) and the number of $k$ indicating the top experts. On \textit{x}-axis, $(n,k)$ is reported, where $n$ corresponds to the number of experts, where $k$ corresponds to the number of the most relevant experts as described in Eq.~\ref{topk_equations}. On $y$-axis, Accuracy is reported. As one can easily observe, in most cases by decreasing the number of experts, Accuracy presents also a decrease. Specifically, $(64,32)$ corresponds to an Accuracy of 0.99875, $(64,16)$ corresponds to an Accuracy of 0.99863, $(32,16)$ corresponds to an Accuracy of 0.99881, $(32,4)$ corresponds to an Accuracy of 0.99862, and $(16,4)$ corresponds to an Accuracy of 0.99834. Therefore, having more experts leads to better performance for our task.

\paragraph{NANCY}

Firstly, we set $L_{load}=L_{importance}=0$. Our proposed model (Table~\ref{results_ourapproach_nancy}) achieved an overall accuracy of 0.79592, whereas this model (Table~\ref{Ablation_Study (NANCY)}) shows a slightly lower accuracy of 0.78614. While the difference in overall accuracy is minor, a closer examination of class-wise performance reveals more insightful distinctions. For the Benign class, the first model slightly outperforms the second, with an F1 score of 0.98679 compared to 0.98389, indicating marginally better consistency in identifying normal traffic. In the case of SYN Scan detection, the first model again shows superior performance across all metrics: precision (0.59807 vs. 0.57368), recall (0.39572 vs. 0.35541), and F1 score (0.47629 vs. 0.43890). Although both models struggle with detecting SYN scans, the first model demonstrates a relatively better capacity to identify this type of stealthy intrusion. For TCP Scan (Connect Scan), the second model performs almost identically to the first in terms of recall (0.74533 vs. 0.74404) but has slightly lower precision (0.52975 vs. 0.64581) and F1 score (0.61932 vs. 0.62969). This suggests that while both models detect similar quantities of true positives, the first model does so with fewer false positives, making it more reliable overall for this class. Regarding SYN flood detection, the F1 scores are nearly identical (0.99070 vs. 0.99162), indicating both approaches are highly effective for this attack type with negligible differences. Finally, for HTTP flood detection, one can observe a drop in F1-score from 0.99774 to 0.99718.

Next, we remove the MoE layer and observe differences in performance. Our proposed model demonstrates better overall accuracy (0.79592 vs. 0.77876). For benign traffic, both models perform similarly, but the proposed model achieves marginally higher precision and F1-score, indicating slightly more reliable identification of normal activity. In detecting SYN Scan attacks, the model without the MoE layer shows a significantly higher recall (0.83337 vs. 0.39572), meaning it captures more true positives, though model with MoE layer maintains better precision, resulting in a higher overall balance in performance. For TCP Scan, the introduced model clearly outperforms model without MoE with much higher recall (0.74404 vs. 0.23221) and F1 score (0.62969 vs. 0.33354), highlighting proposed model’s superior ability to detect this attack type. Both models perform exceptionally well in detecting SYN flood and HTTP flood attacks, with model without MoE layer having slightly better recall in SYN flood detection, though the differences are minimal. Overall, our proposed model incorporating both CNN and MoE provides a more balanced and consistent performance across all traffic classes.

Finally, we use a neural network, i.e., 72 units (input layer), 36 units (hidden layer), 5 units (output layer). Results show that CNN+MoE outperforms the dense-layer model in terms of overall accuracy (0.79592 vs. 0.76558), indicating better general classification performance. For benign traffic, CNN+MoE achieves slightly higher precision, recall, and F1 score, suggesting it identifies normal behavior more accurately. In detecting SYN Scan attacks, the dense-layer model demonstrates notably better recall (0.80135 vs. 0.39572), meaning it detects more true positives, though CNN+MoE maintains higher precision. For TCP Scan detection, both models perform weakly, but CNN+MoE again provides a better balance, with significantly higher recall (0.74404 vs. 0.23689) and F1 score (0.62969 vs. 0.33029). In flood-based attacks, both models perform very well; however, CNN+MoE slightly edges out the dense-layer model in HTTP flood detection, while both are comparably strong for SYN flood. Overall, CNN+MoE exhibits more consistent and balanced performance across all classes, particularly in scan-type attack detection.

\begin{table}[htbp]
\label{Ablation_Study (NANCY)}
\centering
\caption{Ablation Study (NANCY dataset)}
\begin{tabular}{lcccc}
\toprule
\multicolumn{1}{l}{}&\multicolumn{4}{c}{\textbf{Evaluation metrics}}\\
\cline{2-5} 
\multicolumn{1}{l}{\textbf{Architecture}}&\textbf{Accuracy}&\textbf{Precision}&\textbf{Recall}&\textbf{F1-score}\\
\midrule
\multicolumn{5}{>{\columncolor[gray]{.8}}l}{\textbf{Set importance and load loss to zero}} \\
Benign                  & \multirow{5}{*}{0.78614} & 0.99486 & 0.97317 & 0.98389 \\ 
SYN Scan                &                  & 0.57368 & 0.35541 & 0.43890 \\ 
TCP Scan        &                  & 0.52975 & 0.74533 & 0.61932 \\ 
SYN flood               &                  & 0.99754 & 0.98395 & 0.99070 \\ 
HTTP flood              &                  & 0.99802 & 0.99635 & 0.99718 \\ 
\midrule
\multicolumn{5}{>{\columncolor[gray]{.8}}l}{\textbf{Remove MoE layer}} \\
Benign                  & \multirow{5}{*}{0.77876} & 0.99355 & 0.97815 & 0.98579 \\ 
SYN Scan                &                  & 0.51127 & 0.83337 & 0.63374 \\ 
TCP Scan        &                  & 0.59182 & 0.23221 & 0.33354 \\ 
SYN flood               &                  & 0.99496 & 0.98662 & 0.99077 \\ 
HTTP flood              &                  & 0.99817 & 0.99589 & 0.99703 \\ 
\midrule
\multicolumn{5}{>{\columncolor[gray]{.8}}l}{\textbf{Simple input and dense layer}} \\
Benign                  & \multirow{5}{*}{0.76558} & 0.97819 & 0.95929 & 0.96865 \\ 
SYN Scan                &                  & 0.49875 & 0.80135 & 0.61484 \\ 
TCP Scan        &                  & 0.54528 & 0.23689 & 0.33029 \\ 
SYN flood               &                  & 0.99721 & 0.97579 & 0.98638 \\ 
HTTP flood              &                  & 0.98796 & 0.98394 & 0.98595 \\ 
\bottomrule
\end{tabular}
\end{table}

\section{Conclusion and Future Work} \label{sec_conclusion_future_work}

In this paper, we present the first study utilizing Mixture of Experts in the intrusion detection task Specifically, we use a publicly available dataset (5G-NIDD), which is generated from a real 5G test network. Input 1D array of features is transformed into a 2D matrix and then is given as input to CNN layers. The representation vector of the CNN layers is passed through the Mixture of Experts layer consisting of experts and a router. Results on the 5G-NIDD dataset showed that our proposed model outperforms state-of-the-art results reaching Accuracy and F1-score up to 0.99958 and 0.99894 respectively. Results on the NANCY dataset indicate that the proposed approach reaches Accuracy up to 0.78614. Results of an ablation study demonstrated the effectiveness of all the components of our introduced approach. 

\textbf{Limitations:} Our work includes some limitations. Specifically, we used simple methods for imputing missing values, which may influence the evaluation performance. More advanced methods, including Generative Adversarial Imputation Networks (GAIN) \cite{pmlr-v80-yoon18a}, have been proposed throughout the years. Additionally, our study relies on labelled datasets. 

\textbf{Future Work:} In the future, we aim to use advanced missing value imputation methods and combine self - supervised learning with mixture of experts strategies.

\section*{Acknowledgments}

This work is supported by the Smart Networks and Services Joint Undertaking (SNS JU) under the European Union’s Horizon Europe research and innovation programme under Grant Agreement No 101139198, iTrust6G project. Views
and opinions expressed are however those of the author(s) only and do not necessarily reflect those of the European Union or SNS-JU. Neither the European Union nor the granting authority can be held responsible for them.

\bibliographystyle{unsrt}  
\bibliography{references} 

\end{document}